\documentclass{article}

 \usepackage[final,nonatbib]{nips_2018}

\usepackage[utf8]{inputenc} 
\usepackage[T1]{fontenc}    
\usepackage{hyperref}       
\usepackage{url}            
\usepackage{booktabs}       
\usepackage{amsfonts}       
\usepackage{nicefrac}       
\usepackage{microtype}      
\usepackage{amsmath}
\usepackage{tikz}
\usetikzlibrary{arrows,automata}
\usepackage[toc,page]{appendix}

\title{A Deep Latent-Variable Model Application to Select Treatment Intensity in Survival Analysis}

\author{
  C\'edric Beaulac \\
  Department of Statistical Sciences\\
  University of Toronto\\
   \And
   Jeffrey S. Rosenthal \\
  Department of Statistical Sciences\\
  University of Toronto\\
     \And
   David Hodgson \\
   Department of Radiation Oncology\\
  University of Toronto\\
}

\begin{document}


\maketitle

\begin{abstract}
In the following short article we adapt a new and popular machine learning model for inference on medical data sets. Our method is based on the Variational AutoEncoder (VAE) framework that we adapt to survival analysis on small data sets with missing values. In our model, the true health status appears as a set of latent variables that affects the observed covariates and the survival chances. We show that this flexible model allows insightful decision-making using a predicted distribution and outperforms a classic survival analysis model.
\end{abstract}

\section{Introduction}
\label{intro}

Understanding the effect of a treatment $t$ on a response $y$ for an individual patient with characteristics (covariates) $\mathbf{x}$, is a central problem of data analysis in various fields. In medical sciences, it often arises when physicians are working on a treatment to cure a
certain disease. Our collaborators at the Children Oncology Group (COG) provided us with a real data set of children with Hodgkin Lymphoma.  Our goal is to construct an interpretable classifier that selects the right treatment for the right patient.  Since the human body is complicated and there might exist some extremely complex interactions between the characteristics, the treatment and the response of the patient, we need a model that allows for interactions of high order between variables.  Since it is impossible to truly know the degree to which the patient is sick, the patient characteristics $\mathbf{x}$ serve to estimate the true patient health status.  Since the treatments were selected by doctors based on the observed covariates, the data suffers from treatment selection bias \cite{stukel07,Wallis18} that needs to be accounted for.

\smallskip 

To accomplish this, we use a deep-latent variable model inspired by Louizos et al. \cite{Louizos17}.  To account for treatment selection bias, the true patient status is represented by a set of hidden variables $\mathbf{z}$ that affects both the observed characteristics $\mathbf{x}$ and the survival chances $y$ of the patient. Although learning the exact posterior distribution is intractable in many cases \cite{Bishop07,Kingma17}, variational inference allows for inference on latent variable models \cite{Kingma13,Rezende14}. The Variational AutoEncoder (VAE) framework proposed by Kingma \cite{Kingma13,Kingma17} offers interesting properties: the probabilistic modelling allows for interpretable results and useful statistical properties; the neural network parameterization allows for complex relationships between variables as needed; and the latent variables allow for more flexible observed data distributions. Everything is combined in a system that can be jointly optimized using variational inference.

\section{Data set challenges} \label{chal}

Friedman et al. \cite{Friedman14} previously introduced the data set we received. It is a small data set by machine learning standards, the response variable is right-censored for many observations and it contains observations with missing values. VAEs were not designed to handle such data sets and therefore we had to adapt our VAE model in order to face those challenges.

\smallskip

First, since obtaining medical data is expensive, our data set is small. In that situation, underfitting is a concern, but as reproducibility is a major concern in the medical community we must take precautions so that we don't overfit either. To prevent overfitting we relied on L2 regularization, also known as weight decay, and we performed thorough model checking in order to ensure our model is not overfitting. 

\smallskip 

Second, as it is the case in many survival data sets, the response is right-censored for many observations. Therefore, we established a model that is inspired by survival analysis theory and we used classic survival analysis techniques to account for the right-censored observations as will be explained in section \ref{dis}. 

\smallskip

Finally, the data set contains many missing values. Our collaborators indicate that missing values are assumed to be missing at random (MAR) and thus we performed missing values imputation \cite{vanBuuren11,Azur11} during the data processing phase.

\section{The model} \label{model}

The main purpose of latent variables is to allow for more flexible and complicated distributions for the observed variables \cite{Bishop07,Koller09}. Here is a graphical representation of the model we suggest :

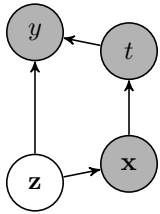
\begin{figure}[h]
\begin{center}
\begin{tikzpicture}[->,>=stealth', semithick]
  \tikzstyle{latent}=[fill=white,draw=black,text=black,style=circle,minimum size=.75cm]
  \tikzstyle{observed}=[fill=black!30,draw=black,text=black,style=circle,minimum size=.75cm]

  \node[latent] 		(A)	at (0,0)		{$\mathbf{z}$};
  \node[observed]         (B)	at (1.25,0.25)		{$\mathbf{x}$};
  \node[observed]         (C)	at (0,2)		{$y$};
  \node[observed]         (D)	at (1.25,1.75)		{$t$};

  \path (A) edge              node {} (B)
            edge              node {} (C);
  \path (B) edge              node {} (D);           
  \path (D) edge              node {} (C);

\end{tikzpicture}
\end{center} 
\caption{The graphical representation of our deep-latent variable model. The response is identified by $y$, the treatment by $t$, the observed characteristics by $\mathbf{x}$ and the patient health status by $\mathbf{z}$.}
\label{graph}
\end{figure}

The representation illustrated in figure \ref{graph} induces a natural factorization of the joint distribution. As said earlier, from a practical perspective, the latent variables $\mathbf{z}$ are incorporated to allow for a more flexible observed data distribution but the graphical model also leads to an intuitive description. Here, the set of latent variables represents the true patient health status and directly affects the survival chances of the patient $y$ and the covariates $\mathbf{x}$ gathered as proxy of the true health status. The treatment $t$ is considered a special covariate as it is selected by a physician based upon the observed covariates $\mathbf{x}$. Finally, the distribution for the response $y$ is based upon the patient health status $\mathbf{z}$ and the treatment selected $t$. Neural network parameterizations along edges of the graphical representation insure that the model includes high order of interactions between the variables which is important to us but difficult to do with the Cox PH model.

\subsection{Model distributions}
\label{dis}

This model illustrated in figure \ref{graph} suggest the following factorization :

\begin{equation}
p_{\theta}(\mathbf{z},\mathbf{x},t,y) = p_\theta(\mathbf{z})p_\theta(\mathbf{x}|\mathbf{z})p_\theta(t | \mathbf{x}) p_\theta(y|t,\mathbf{z}).
\label{fact}
\end{equation}

The prior $p_\theta(\mathbf{z})$ is a multivariate Normal distribution with mean 0 and identity variance. As the graphical representation suggests, the observed covariates $\mathbf{x}$ are conditionally independent given the latent variables $\mathbf{z}$. The conditional distributions $p_\theta(\mathbf{x} | \mathbf{z})$ can take multiple forms such as Normal, Poisson, Bernoulli, etc. We want to understand the effect of additional treatments such has intensive chemotherapy or radiotherapy. Within the collected data set, these treatments are either received or not by the patients and therefore we've established them as a set of Bernoulli variables. 

\smallskip

Finally, the response is modelled as Weibull. In order to account for censored data the log-likelihood for the survival distribution will be computed as follows :

\begin{equation}
\log p_{\theta}(y | t,\mathbf{z}) = \delta \log f_{\theta}(y | t,\mathbf{z}) + (1-\delta) \log S_{\theta}(y |t,\mathbf{z}),
\label{surv}
\end{equation}

where $\delta = 1$ if $y$ is observed and 0 if $y$ is censored, $f_\theta$ is the density and $S_{\theta}$ is the survival function. More information about distributions and parameterizations is located in the appendices.

\subsection{Fitting the parameters}

The parameters of the various neural networks will require training. The Evidence Lower BOund (ELBO) is a lower bound for the log-likelihood of the observed data and will be the objective function to maximize during training. Using the factorization (\ref{fact}) explained in section \ref{dis} we have :

\begin{equation}
\begin{split}
\text{ELBO}  &=  \mathbf{E}_{q_\phi} \left[ \ln \frac{  p_\theta(\mathbf{x},t,y,\mathbf{z})}{ q_\phi(\mathbf{z}|\mathbf{x},y)} \right] =  \mathbf{E}_{q_\phi} \left[ \ln  p_\theta(\mathbf{x},t,y,\mathbf{z}) -\ln q_\phi(\mathbf{z}|\mathbf{x},y) \right] \\ 
&=  \mathbf{E}_{q_\phi}\left[ \text{ln} p_\theta(\mathbf{z}) + \text{ln} p_\theta(\mathbf{x} | \mathbf{z}) + \text{ln} p_\theta(t|\mathbf{x}) + \text{ln} p_\theta(y |t,\mathbf{z}) - \text{ln} q_\phi(\mathbf{z}|\mathbf{x},y) \right].
\end{split}
\label{ELBO}
\end{equation}

Since the observed data parameters $\theta$ and the variational distribution parameters $\phi$ are obtained through various neural network functions, we will attempt to maximize the ELBO with respect to the neural network functions parameters. This will require the use of back-propagation combined with a gradient-based optimizer. 

\subsection{Prediction}

The ultimate goal of this analysis is to provide tools to physicians to allow them to make a decision about the treatment needed for a patient. With our probabilistic approach we aim at giving physicians a wide range of information which they can utilize however they see fit. Our model produces a predicted distribution for the event-free survival time of a patient based upon its characteristics and the selected treatment. Having such distributions for every possible treatment gives flexibility regarding decision-making as physicians can look at various properties of the predicted distributions such as the expected value or the survival function.  Using importance sampling our model let us produce the following predicted distribution for a new patient with characteristics $\mathbf{x}$ for a given treatment $t$ :

\begin{align}
\label{ISy}
p(y|t,\mathbf{x}) = \sum_{l=1}^L w_l p_\theta(y|t,\mathbf{z}_l)
\end{align}

which resembles a mixture of Weibull where $L$ is the number of components, and $w_l$ is the component weight. The appendices contain more details about the importance sampling formulation.

\section{Results}

\begin{figure*}[ht]
\begin{center}
\includegraphics[width=6.75cm,height=5cm]{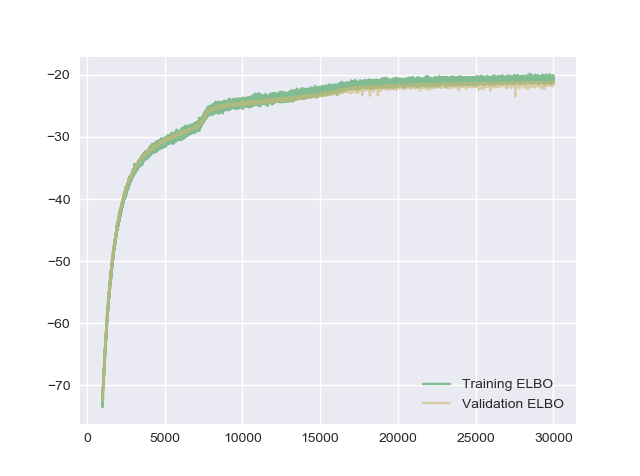}
\includegraphics[width=6.75cm,height=5cm]{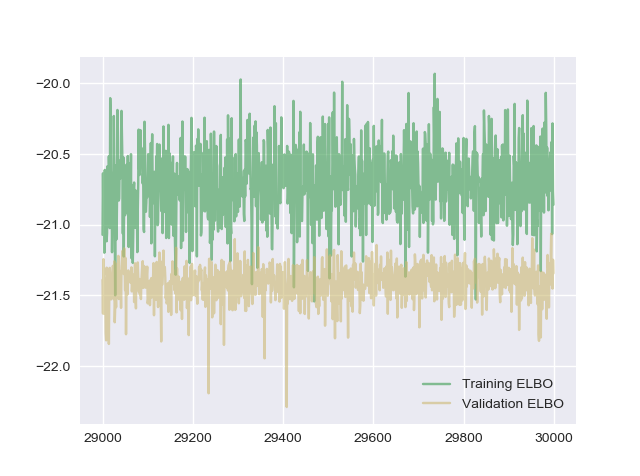}
\caption{Log-likelihood lower bound (ELBO) through out epochs. We stopped the optimization procedure when the training set ELBO stabilized and before the validation set ELBO decreased. }
\label{elbo} 
\end{center}
\end{figure*}

We used the architecture presented in section \ref{model} and optimized the parameters using the ADAM optimizer \cite{Kingma14,Kingma17}. Model selection and calibration is performed using a validation set. Since the log-likelihood estimates on a held-out set are a better estimates of the log-likelihood under the true data generation distribution \cite{Shai14} selecting the tuning parameters using the validation set contributes towards preventing overfitting. Similarly the gap between the log-likelihood on the training set and the validation set is small in proportion to the log-likelihood estimates on the validation set in figure \ref{elbo}. This indicates that we have managed to prevent overfitting which was one of the challenges we established in section \ref{chal}.

\smallskip  

Because of the censored observations, the accuracy of our model cannot be assessed with the mean-squared error. The concordance index (\textit{c}-index) \cite{Harrell96} is one of the most popular performance measures for censored data \cite{Chen12,Steck07}. It is a rank-based measure that computes the proportion of all usable observation pairs in which the predictions and true outcomes are concordant \cite{Harrell96}. A \textit{c}-index of 0.5 is equivalent to random ordering and 1 is perfect ordering. 

\smallskip

The concordance index on the training set is 0.682 for the Cox PH model against 0.649 for our model, but on the validation set this index is equal to 0.522  for the predicted hazards using the Cox PH model and 0.574 for the VAE we propose. This indicates that our model suffers less from overfitting than Cox PH and outperforms it on held-out data according to this performance measure. 

\smallskip

Finally, let's quickly introduce an example of decision-making. With our model, the predicted survival function can be obtained quite simply : 

\begin{equation}
\begin{split}
P(Y > y|t,\mathbf{x})  = \sum_{l=1}^L w_l  P_\theta(Y > y|t,\mathbf{z}_l).  \\
\end{split}
\label{ps}
\end{equation}

The predicted survival function of equation \ref{ps} can be used to estimate the increase in survival chances caused by selecting treatment $t_1$ instead of treatment $t_0$ by computing : $ P(Y > y|t_1,\mathbf{x})  -  P(Y > y|t_0,\mathbf{x})$. For example, our collaborators' suggested decision-making was to intensify the treatment using either intensive chemotherapy or radiation therapy if it increases the 4 year event-free survival chances by at least 7\%. Our model would allow to make such decisions.

\section{Conclusion}

We have adapted a new machine learning technique to a typical medical framework and compared the results to a common technique. Three challenges (small data set, censored data, and
missing values) were identified and we managed to adapt our model in order to face them. The result is a VAE adapted for survival analysis that allows for complex interactions between the variables, that accounts for treatment selections bias, that can generate a predicted distribution for a new patient allowing for insightful decision-making and that outperforms the popular Cox PH model in terms of prediction accuracy.   

\smallskip

As future improvements we are looking at alternatives to face the challenges identified in section \ref{chal}. We would like to establish a recognition model that accounts for missing data such as recommended by Nazabal et al. \cite{Nazabal18}. We would also like to utilize other regularization techniques to prevent overfitting, such as dropout \cite{Hinton12,Srivastava14} and variational dropout \cite{Kingma17,Kingma15}. Finally, we would like to test the accuracy of our model with other performance measures and to compare our model with a wider range of classic survival models.

\section*{Acknowledgement}
The authors are grateful to Qinglin Pei and the rest of the Children Oncology Group (COG) staff for collecting, handling and providing us with this data set. The authors are thankful to David Duvenaud and Chris Cremer for their insightful guidance. Finally, the authors would like to acknowledge the financial support of the NSERC of Canada.


\bibliographystyle{plain} 
\bibliography{mybibfile}   

\clearpage

\begin{appendices}
\label{append}

\subsection{Model distributions}

In order to maximize the ELBO as establish in equation \ref{ELBO}, we need to establish the joint distribution. Based upon the factorization suggested in figure \ref{graph}, the joint distribution can be expressed as :

\begin{equation}
p_{\theta}(\mathbf{z},\mathbf{x},t,y) = p_\theta(\mathbf{z})p_\theta(\mathbf{x}|\mathbf{z})p_\theta(t | \mathbf{x}) p_\theta(y|t,\mathbf{z}).
\end{equation}

We have decided to set the prior distribution of the latent variables to a simple Normal ball :

\begin{equation}
p(\mathbf{z}) =  \mathcal{N}(\mathbf{z} | 0,I).
\end{equation}

The size of the latent space can be considered a tuning parameters, using the validation set we decided upon a latent space of dimension 4. As the graphical representation suggests, the observed predictors $\mathbf{x}$ are conditionally independent given the latent variables $\mathbf{z}$ :

\begin{equation}
p_\theta(\mathbf{x} | \mathbf{z}) = \prod_{j=1}^{D_x} p_\theta(x_{j} | \mathbf{z})	
\end{equation}

In our model, $t$ represents possible additional treatments; intensive chemotherapy and radiation therapy. Both of these are separately either given or not, thus a Bernoulli distribution is well suited to model these two variables :

\begin{equation}
p(t_i | \mathbf{x}) = \text{Ber}(\hat{\pi}_i) \text{ for } i \in \{1,2\}.
\end{equation}

Finally, for our first attempt at analysing this data set, the distribution of the response $y$ was set to be Weibull, a common distribution in the survival analysis literature  :
 
\begin{equation}
p(y | t, \mathbf{z}) = \text{Weibull}(\lambda,K)
\end{equation}

In order to allow for interactions between variables, for a complex relationship between the latent variables and the observed ones and for a general set up that might get expanded to more applications, we have utilized a neural network parameterization. More specifically, we use neural networks along all edges of the graph in figure \ref{graph} to represent the relationship between the set of parent variables and the parameters of the children distributions. Explicitly :

\begin{align}
\theta &= f_2 (\mathbf{W}_2 f_1(\mathbf{W}_1\mathbf{z} + \mathbf{b}_1) + \mathbf{b}_2) \\
 [\pi_1,\pi_2 ] &= f_4 (\mathbf{W}_4 f_3(\mathbf{W}_3\mathbf{x} + \mathbf{b}_3) + \mathbf{b}_4) \\
  [\lambda,K] &= f_6 (\mathbf{W}_6 f_5(\mathbf{W}_5[\mathbf{z},t] + \mathbf{b}_5) + \mathbf{b}_6),
\label{decodp}
\end{align}

where $f_i$ are activation functions, $\mathbf{W}_i$ are matrices of weights and $\mathbf{b}_i$ are vectors of biases. Finally, we have to define a variational distribution, which is an approximation of the true posterior of $\mathbf{z}$ given the observed variables. In our experiments, we used the following variational distribution :

\begin{equation}
q( \mathbf{z} | \mathbf{x},y) = \mathcal{N}(\mathbf{z} | \mathbf{\mu} ,\sigma^2I)
\end{equation}

where the parameterization is established again with a neural network :

\begin{align}
 [\mu,\sigma ] &= f_8 (\mathbf{W}_8 f_7(\mathbf{W}_7[\mathbf{x},y] + \mathbf{b}_7) + \mathbf{b}_8). 
 \label{encordp}
\end{align}

$\mathbf{W}_i$ and $\mathbf{b}_i$ for $i \in \{1,...,8\}$ are the parameters that require training.

\subsection{Evidence Lower Bound}

Here is a quick decomposition of the observed-data log-likelihood that explains why the ELBO is a lower bound :

\begin{equation}
\begin{split}
\ln p(\mathbf{x},t,y) &=  \mathbf{E}_{q(\mathbf{z}|\mathbf{x},y)} \left[ \ln p(\mathbf{x},t,y) \right] \\
&=  \mathbf{E}_{q(\mathbf{z}|\mathbf{x},y)} \left[ \ln \frac{ p(\mathbf{x},t,y,\mathbf{z})}{p(\mathbf{z}|\mathbf{x},y,t)} \right] \\
&=  \mathbf{E}_{q(\mathbf{z}|\mathbf{x},y)} \left[ \ln \frac{ p(\mathbf{x},t,y,\mathbf{z})q(\mathbf{z}|\mathbf{x},t)}{q(\mathbf{z}|\mathbf{x},y)p(\mathbf{z}|\mathbf{x},y,t)} \right] \\
&=  \mathbf{E}_{q(\mathbf{z}|\mathbf{x},y)} \left[ \ln \frac{ p(\mathbf{x},t,y,\mathbf{z})}{ q(\mathbf{z}|\mathbf{x},y)} \right] + \mathbf{E}_{q(\mathbf{z}|\mathbf{x},y)} \left[ \ln \frac{ q(\mathbf{z}|\mathbf{x},y)}{p(\mathbf{z}|\mathbf{x},y,t)} \right] \\
&= \mathcal{L}(q,\theta) + KL(q||p) \\
&\geq \mathcal{L}(q,\theta) 
\end{split}
\end{equation}

where  $\mathcal{L}(q,\theta)$ is the ELBO. 

\subsection{Importance sampling for predictions}

Under the parameterization of equation \ref{fact} induced by the graphic of figure \ref{graph} we cannot simply produce an estimate for $p(y|t,\mathbf{x})$, the density for the response given the treatment and the patient characteristics. Thus we will need to rely on an importance sampling technique as follows :

\begin{equation}
\begin{split}
p(y|t,\mathbf{x}) &= \int_\mathbf{z} p(y|t,\mathbf{x},\mathbf{z}) p(\mathbf{z}|t,\mathbf{x}) d\mathbf{z} \\
 &= \int_\mathbf{z} p_\theta(y|t,\mathbf{z}) p(\mathbf{z}|t,\mathbf{x}) d\mathbf{z} \\
\end{split}
\end{equation}

Since we cannot sample directly from $p(\mathbf{z}|t,\mathbf{x})$ we need to find a distribution of $\mathbf{z}$ from which we can easily sample. The prior $p_\theta(\mathbf{z})$ is easy to sample from, thus :

\begin{equation}
\begin{split}
p(y|t,\mathbf{x}) &= \int_\mathbf{z} p_\theta(y|t,\mathbf{z}) p(\mathbf{z}|t,\mathbf{x}) d\mathbf{z}  \\
  &= \int_\mathbf{z} p_\theta(y|t,\mathbf{z}) \frac{p(\mathbf{z}|t,\mathbf{x})}{ p_\theta(\mathbf{z})} p_\theta(\mathbf{z}) d\mathbf{z} \\
  &\approx \frac{1}{L} \sum_{l=1}^L r_l  p_\theta(y|t,\mathbf{z}_l)
\end{split}
\end{equation}

where $r_l = p(\mathbf{z}_l|t,\mathbf{x})/ p_\theta(\mathbf{z}_l)$. The above will be a mixture of Weibull of $L$ components with weights $r_l/L$. One might notice that we cannot evaluate $p(\mathbf{z}_l|t,\mathbf{x})$ with our current model, but we can up to a normalization constant which leads to the following :

\begin{equation}
\begin{split}
p(y|t,\mathbf{x}) &= \int_\mathbf{z} p_\theta(y|t,\mathbf{z}) \frac{p(\mathbf{z}|t,\mathbf{x})}{p_\theta(\mathbf{z})}p_\theta(\mathbf{z}) d\mathbf{z} \\
&= \int_\mathbf{z} p_\theta(y|t,\mathbf{z}) \frac{p(\mathbf{z},t,\mathbf{x})}{p_\theta(\mathbf{z})p(\mathbf{x},t)} p_\theta(\mathbf{z}) d\mathbf{z} \\
&= \int_\mathbf{z} p_\theta(y|t,\mathbf{z}) \frac{p_\theta(\mathbf{z})p_\theta(\mathbf{x}|\mathbf{z}) p(t|\mathbf{x})}{p_\theta(\mathbf{z})p(\mathbf{x})p(t|\mathbf{x})} p_\theta(\mathbf{z}) d\mathbf{z} \\
&= \int_\mathbf{z} p_\theta(y|t,\mathbf{z}) \frac{p_\theta(\mathbf{x}|\mathbf{z}) }{p(\mathbf{x})} p_\theta(\mathbf{z}) d\mathbf{z} \\
  &\approx \frac{1}{L} \frac{1}{p(\mathbf{x})} \sum_{l=1}^L p_\theta(\mathbf{x}|\mathbf{z})  p_\theta(y|t,\mathbf{z}_l)
\end{split}
\end{equation}

Now we can also use the same samples to evaluate the normalization constant $p(\mathbf{x})$ :

\begin{equation}
\begin{split}
p(\mathbf{x}) = \int_\mathbf{z} p_\theta(\mathbf{x}|\mathbf{z})  p_\theta(\mathbf{z})d\mathbf{z}
  &\approx \frac{1}{L} \sum_{l=1}^L p_\theta(\mathbf{x}|\mathbf{z}_l) 
\end{split}
\end{equation}

both of these results combined lead to :

\begin{equation}
\begin{split}
p(y|t,\mathbf{x})  &\approx  \sum_{l=1}^L w_l  p_\theta(y|t,\mathbf{z}_l)
\end{split}
\end{equation}

where :

\begin{equation}
\begin{split}
w_l = \frac{p_\theta(\mathbf{x}|\mathbf{z}_l)}{\sum_{k=1}^L p_\theta(\mathbf{x}|\mathbf{z}_k)}
\end{split}
\end{equation}

\subsection{Insightful decision-making}

Through out the article we mentioned that a predicted distribution offers more flexibility than point estimation. Here, we will mention a few examples of information that can be extracted from the predicted distribution.

\smallskip

To begin, we could easily compute the expected survival time :

\begin{equation}
\begin{split}
\mathbf{E}[y|t,\mathbf{x}]  &= \mathbf{E}\sum_{l=1}^L w_l  p_\theta(y|t,\mathbf{z}_l)  \\
 &= \sum_{l=1}^L w_l  \mathbf{E}(y|t,\mathbf{z}_l).
\end{split}
\end{equation}

Survival function can also be obtained quite simply :

\begin{equation}
\begin{split}
P(Y > y|t,\mathbf{x})  = \sum_{l=1}^L w_l  P_\theta(Y > y|t,\mathbf{z}_l).  \\
\end{split}
\end{equation}

An advantage of our proposed model is that it allows for different decision-making; a physician could be interested in three years survival chances, another physician might prefer to estimate four years survival chances and one might be interested in the expected survival. As mentioned in the main text, we could also easily estimate the increase in survival chances for a given time $y$ by selecting treatment $t_1$ instead of treatment $t_0$ by computing :

\begin{equation}
\begin{split}
P(Y > y|t_1,\mathbf{x})  -  P(Y > y|t_0,\mathbf{x}). \\
\end{split}
\end{equation}

Similarly, interactions between some patient characteristics and treatments could be observed with such models. 

\end{appendices}

\end{document}